\documentclass[sigconf]{acmart}
\usepackage{booktabs} 
\pdfoutput=1
\newcommand{\eg} {\emph{e.g.}}

\newcommand{\etal} {\emph{et al.}}
\newcommand{\ie} {\emph{i.e.}}

\newcommand{\tabincell}[2]{\begin{tabular}{@{}#1@{}}#2\end{tabular}}

\copyrightyear{2018} 
\acmYear{2018} 
\setcopyright{acmcopyright}
\acmConference[MM '18]{2018 ACM Multimedia Conference}{October 22--26, 2018}{Seoul, Republic of Korea}
\acmBooktitle{2018 ACM Multimedia Conference (MM '18), October 22--26, 2018, Seoul, Republic of Korea}
\acmPrice{15.00}
\acmDOI{10.1145/3240508.3240534}
\acmISBN{978-1-4503-5665-7/18/10}

\begin{document}

\fancyhead{}

\title{Temporal Sequence Distillation: Towards Few-Frame Action Recognition in Videos}

\author{Zhaoyang Zhang}
\affiliation{
\institution{Wuhan University} 
\institution{SenseTime Research}}
\email{zhangzhaoyang@whu.edu.cn}

\author{Zhanghui Kuang}
\authornote{Corresponding author is Zhanghui Kuang.}
\affiliation{
\institution{SenseTime Research}}
\email{kuangzhanghui@sensetime.com}

\author{Ping Luo}
\affiliation{
\institution{The Chinese University of Hong Kong}}
\email{pluo@ie.cuhk.edu.hk}

\author{Litong Feng}
\affiliation{
\institution{SenseTime Research}}
\email{fenglitong@sensetime.com}

\author{Wei Zhang}
\affiliation{
\institution{SenseTime Research}}
\email{wayne.zhang@sensetime.com}

\renewcommand{\shortauthors}{Z.Zhang et al.}

\begin{abstract}

Video Analytics Software as a Service (VA SaaS) has been rapidly growing in recent years. VA SaaS is typically accessed by users using a lightweight client. Because the transmission bandwidth between the client and cloud is usually limited and expensive, it brings great benefits to design cloud video analysis algorithms with a limited data transmission requirement. Although considerable research has been devoted to video analysis, to our best knowledge, little of them has paid attention to the transmission bandwidth limitation in SaaS. As the first attempt in this direction, this work introduces a problem of few-frame action recognition, which aims at maintaining high recognition accuracy, when accessing only a few frames during both training and test. Unlike previous work that processed dense frames, we present Temporal Sequence Distillation (TSD), which distills a long video sequence into a very short one for transmission. By end-to-end training with 3D CNNs for video action recognition, TSD learns a compact and discriminative temporal and spatial representation of video frames. On Kinetics dataset, TSD+I3D typically requires only 50\% of the number of frames compared to I3D~\cite{carreira2017quo}, a state-of-the-art video action recognition algorithm, to achieve almost the same accuracies. The proposed TSD has three appealing advantages. Firstly, TSD has a lightweight architecture, and can be deployed in the client, \eg, mobile devices, to produce compressed representative frames to save transmission bandwidth. Secondly, TSD significantly reduces the computations to run video action recognition with compressed frames on the cloud, while maintaining high recognition accuracies. Thirdly, TSD can be plugged in as a preprocessing module of any existing 3D CNNs. Extensive experiments show the effectiveness and characteristics of TSD.

\end{abstract}



\maketitle

\section{Introduction}
\begin{table*}[t]
 \centering
 \small
 \caption{Comparison of methods in VA SaaS scenarios.  \#clips and \#params indicate the  clip number and the parameter number of the model during evaluation. State-of-art methods I3D, TSN, and Res18 ARTNet could only be deployed on cloud, so we report their time of sampling and cropping clips as client runtime. For I3D+TSD, the client runtime consists of both sampling/cropping time and extra computations of TSD. We can see that the I3D+TSD model achieves an accuracy of 72.4, with 238G total FLOPS and only 120 frames required to be transmitted. When efficiency is given precedence,  I3D+TSD achieves an accuracy of 70.7, which is comparable to the I3D baseline. But it needs only 60 frames to be transmitted rather than 250 frames.}
 \begin{tabular}{l|c|c|c|c|c|c|c|c|c|c|c}
\hline
& \multicolumn{5}{c|}{client} & \multicolumn{1}{c|}{transmission} & \multicolumn{5}{c}{server}\\
\cline{2-12} & \tabincell{c}{$\#$clips}
  & \tabincell{c}{$\#$params\\(M)} & \tabincell{c}{FLOPS\\(G)} & \tabincell{c}{runtime\\(ms)} & \tabincell{c}{processed\\$\#$frames} & \tabincell{c}{$\#$frames} &
  \tabincell{c}{backbone} & \tabincell{c}{$\#$params\\(M)} & \tabincell{c}{FLOPS\\(G)} & \tabincell{c}{runtime\\(ms)} & \tabincell{c}{accuracy}\\
 \hline
 I3D \cite{carreira2017quo} & 1 & - & - & 1 & 250 (1$\times$250) & 250 (1$\times$250) & Inceptionv1 & 12 & 422 & 327 & 71.1 \\
 TSN \cite{wang2016temporal} & 25 & - & - & 1 & 250 (25$\times$10) & 250 (25$\times$10) & Inceptionv2 & 10 & 508 & 399  & 69.1 \\
 Res18 ARTNet \cite{wang2017appearance} & 25 & - &- & 4 & 4000 (25$\times$160) & 4000 (25$\times$160) & Resnet18 & 33 & 4850 & 3691 & 69.2 \\

 \hline
 I3D+TSD & 3 & 4 & 36 & 33 & 240 (3$\times$80) & \textbf{60} (3$\times$20) & Inceptionv1 & 12 & \textbf{101} & \textbf{78} & 70.7  \\
 I3D+TSD & 3 & 4 & 36 & 33 & 240 (3$\times$80) & \textbf {120} (3$\times$40) & Inceptionv1  & 12  & \textbf{202} & \textbf{154} & \textbf{72.4}  \\
 I3D+TSD & 10 & 4 & 120 & 109 & 800 (10$\times$80) & 200 (10$\times$20) & Inceptionv1 & 12 & 338  & 258 & 72.1  \\
 I3D+TSD & 10 & 4 & 120 & 109 & 800 (10$\times$80) & 400 (10$\times$40) & Inceptionv1 & 12 & 676 &  511 & \textbf{73.2}  \\

 \hline
 \end{tabular}
 \label{tab:comp}
\end{table*}

\begin{figure}
\centering
\includegraphics[width=1.0\linewidth]{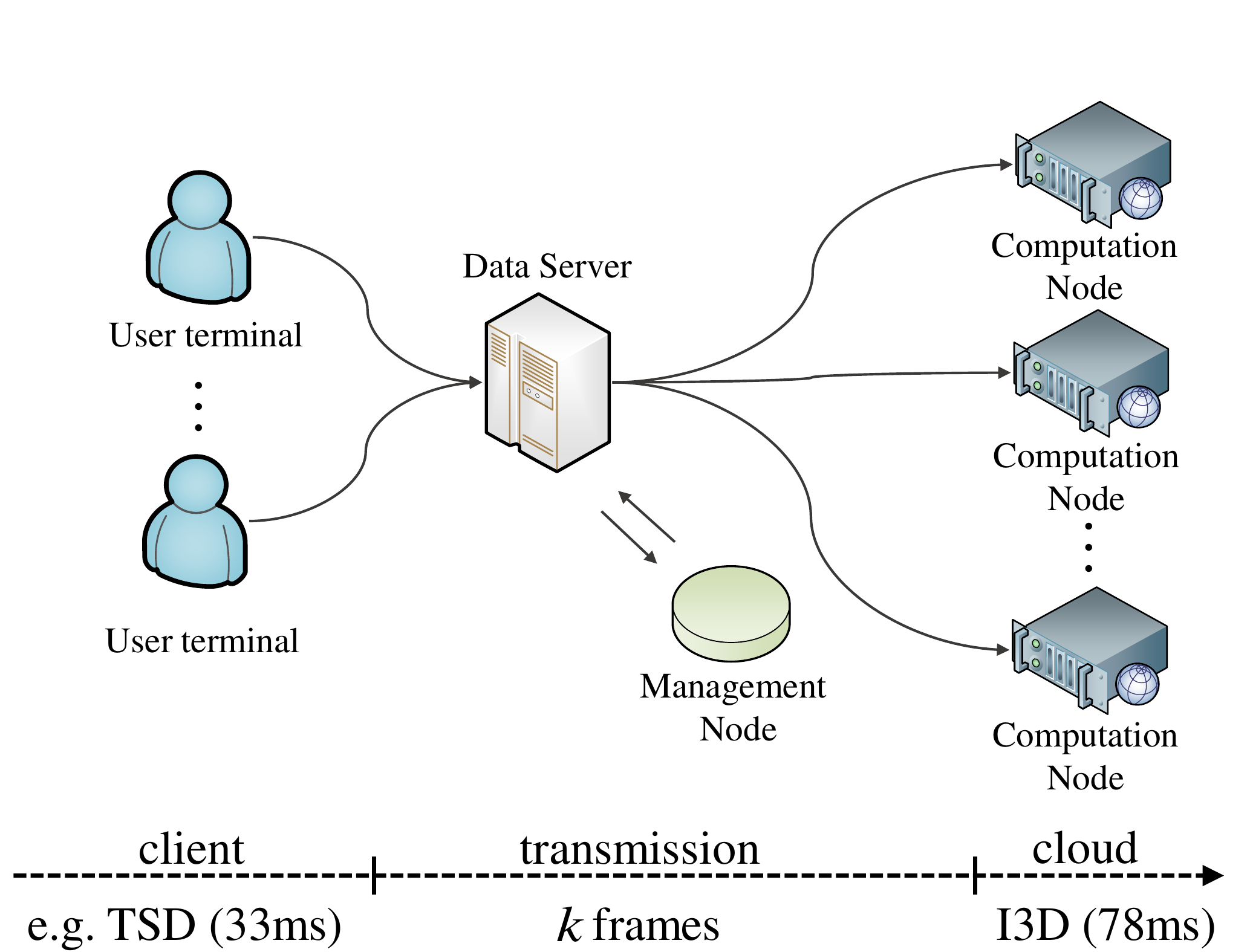}
\caption{Architecture of VA SaaS. It consists of the client-part and the cloud-part. Video frames are transmitted between them. The proposed TSD can be deployed on user terminals to learn compact frame representations. It reduces frame transmission and computations on the cloud and maintains high recognition accuracy. For example, TSD+I3D spends 33ms and 78ms on the client and the cloud respectively, while maintaining a high accuracy of 70.7 on Kinetics when only transmits 60 frames between the client and the cloud. It significantly surpasses the conventional I3D, which transmits 250 frames and takes 327ms on the cloud.}
\label{fig:cloud}
\end{figure}

\begin{figure*}
\centering
\includegraphics[width=1.0\linewidth]{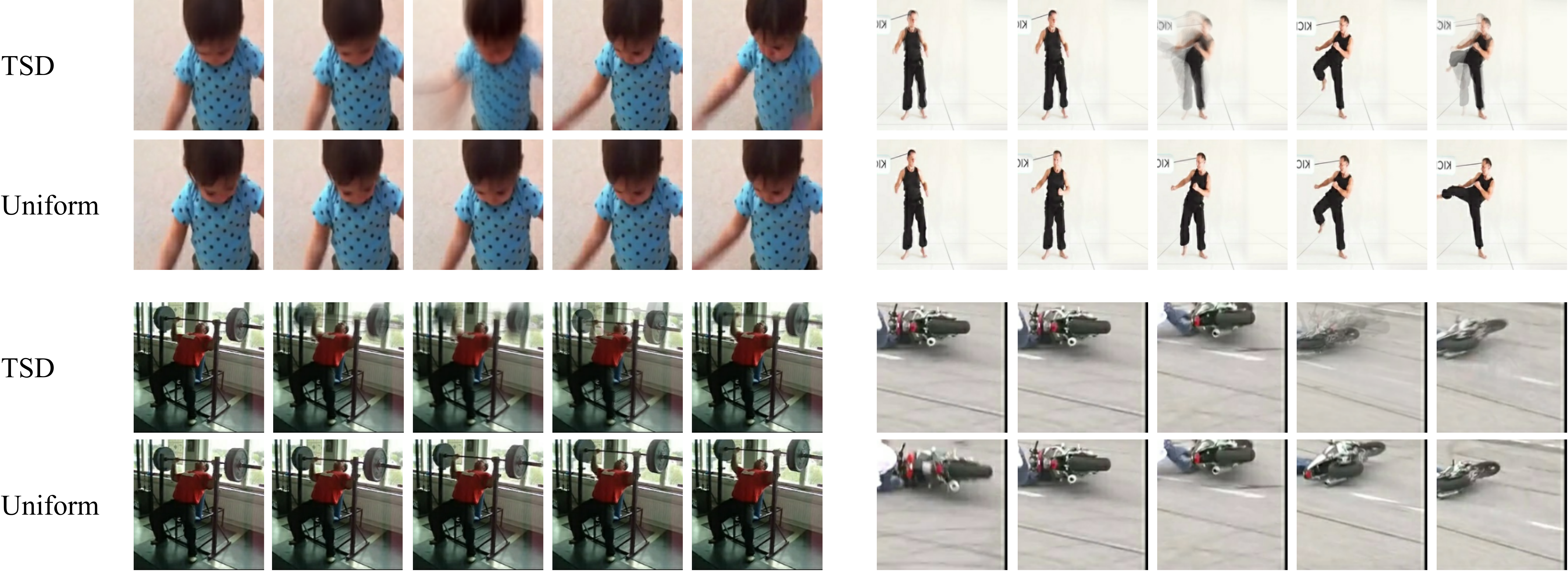}
\caption{Examples of generated frames of TSD and uniform sampling. The first and the third rows show generated frames of TSD while the second and the fourth show the frames uniformly sampled from original RGB sequences.}
\label{fig:sample}
\end{figure*}

Action recognition in videos has large progresses due to the developments of deep Convolution Neural Networks (CNNs)~\cite{simonyan2014very,ioffe2015batch,he2016deep} and the presences
of large-scale video datasets such as Kinetics~\cite{carreira2017quo} and UCF101 ~\cite{soomro2012ucf101}.
Existing work can be generally divided into two categories, 2D CNNs~\cite{wang2016temporal, simonyan2014two} and 3D CNNs~\cite{tran2015learning, carreira2017quo}.

In the first category, 2D CNNs are utilized to extract feature for each frame separately and then pool their predictions across the whole video. For capturing temporal information, 2D CNNs could be adjusted by aggregating features over time~\cite{wang2016temporal}, or fusing multiple modalities like RGB difference and optical flow~\cite{simonyan2014two, wang2016temporal}.
 Although 2D CNNs can be trained by sparsely sampling frames to reduce storages (\eg, GPU memory) and computations, 2D CNNs have limitations in at least two aspects. First, 2D CNNs (\eg, TSN~\cite{wang2016temporal}, which is current state-of-art method) are prone to treat video as an unstructured bag of frames, ignoring the sequential information of video frames. Moreover, for 2D CNNs, the ability to capture long-range dependencies, which is crucial for action recognition performance, is restricted due to the independent frame feature extraction.

In the second category, the above 2D models were extended to 3D models~\cite{tran2015learning}. For instance, I3D~\cite{carreira2017quo} introduced the operations of 3D convolutions to capture long-term temporal information, which is achieved by the expansive receptive field from dense stack of temporal convolution.
I3D achieved state-of-the-art performances by inflating 2D convolutional filters of InceptionNet~\cite{ioffe2015batch} into 3D.
It was trained using 64-frame video clips and tested using the entire video (\eg, 64 frames on UCF101).
Specifically, in training, it randomly samples 64 frames with both spatially and temporal clipping. In test, the network fetches the center clip of the whole video as inputs, averaging predictions temporally.


Although the above approaches achieved good performances for action recognition, their requirements of dense-frame predictions impede their applications in practice, such as Video Analytics Software as a Service (VA SaaS), which has increasing demands in the recent years.
%
Fig.~\ref{fig:cloud} illustrates the architecture of one typical VA SaaS, which has two parts, including the client-part and the cloud-part.
The client-part has multiple terminals such as mobile phones, which have limited memories and computational resources.
The cloud-part contains data server and computation nodes, where the data server distributes data (\eg, video frames) to different computation nodes, while each computation node performs video analysis (\eg, action recognition) and returns results (\eg, predicted action labels) to the data server.
The transmission bandwidths between the client and the cloud, storage resources, and computation resources are typically constrained, in the demands of reducing their cost. For example, a recognition system in the cloud is often permitted to access just a small number of frames but is required to produce results as good as possible.
%


As previous work of action recognition are typically resource consuming, they are not applicable in VA SaaS as summarized in Table~\ref{tab:comp}, which compares different methods in terms of model storages and computations in client and cloud, numbers of frames in transmission and processing, as well as recognition accuracy. The conventional methods~\cite{carreira2017quo, wang2017appearance, wang2016temporal} employ dense-random sampling to select video frames, forming frame clips.
When deploying them in VA SaaS, although each of them has their own merits,
we observe that all of the previous methods are time and transmission bandwidth consuming.



To resolve the above issues, this work presents a new problem of \textbf{few-frame action recognition}, which aims at maintaining high accuracy, but accessing only a few  frames during training and test.
To this end, we propose Temporal Sequence Distillation (TSD), which distills a long video sequence into a short one consisting of compact and informative frames, while preserving temporal and spatial discriminativeness as much as possible.


Fig.~\ref{fig:sample}
illustrates the frames of the distilled sequence of TSD compared with uniformly sampled frames from the original video sequences. The output frames of TSD encode actions with ``ghosting''. In this way, they are compact and informative by reducing the temporal redundancy of videos while distilling action-relevant information.

We design TSD as a lightweight differentiable CNN. It is a general component which can be plugged into any 3D CNNs based action recognition framework.  Since I3D~\cite{carreira2017quo} achieves very impressive results for action recognition, we choose it as our start point to demonstrate the effectiveness of the proposed TSD.

\begin{figure}
\centering
\includegraphics[width=1.0\linewidth]{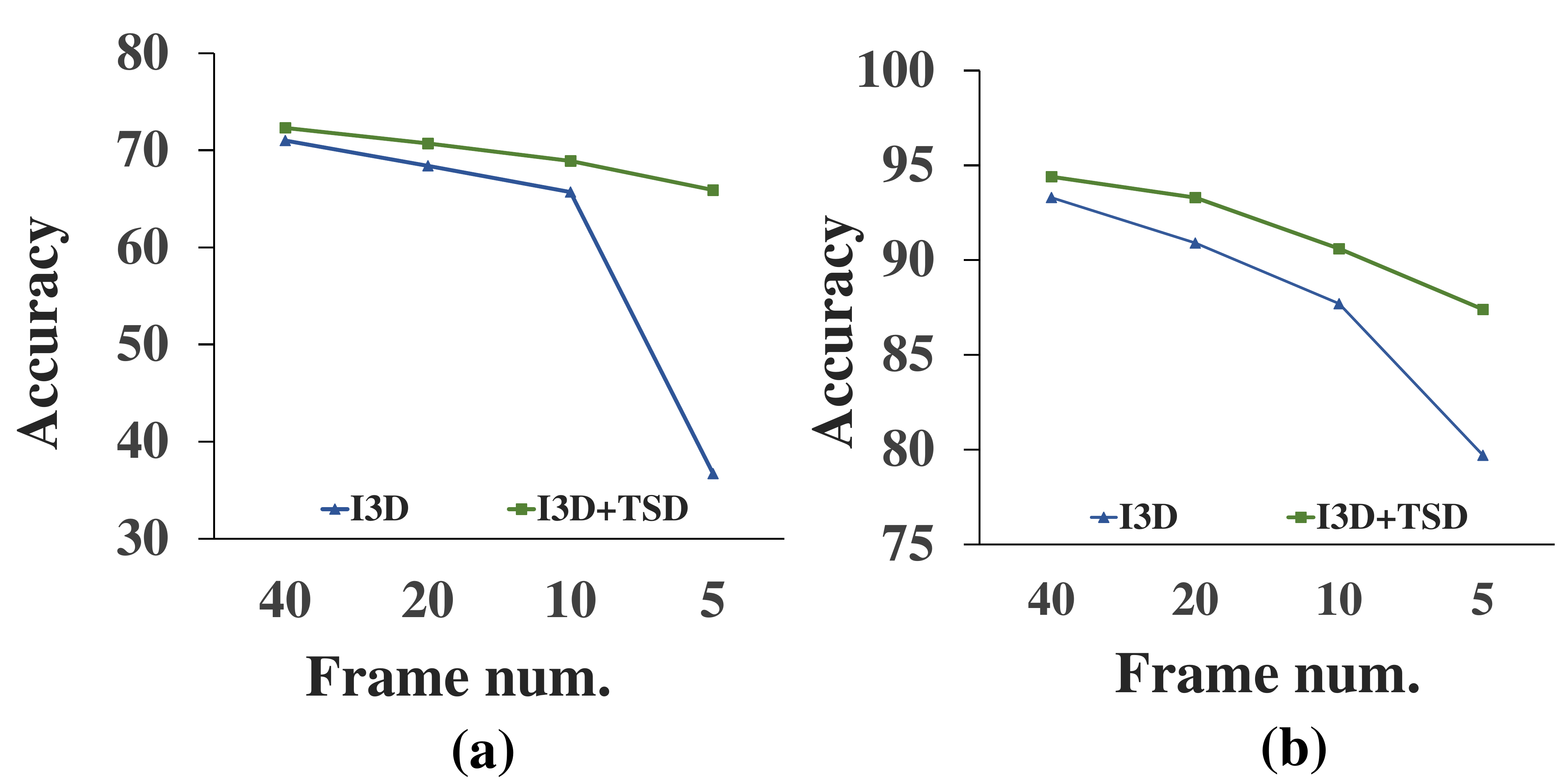}
\caption{Effect of processed frame numbers of I3D and I3D+TSD during testing. (a) and (b) show performance on Kinetics and UCF101 split 1 respectively. More details can be found in Section~\ref{sec:exp}.  }
\label{fig:decay}
\end{figure}

We extensively evaluate TSD enhanced I3D (I3D+TSD) with few frames as input and compare with its counterpart I3D. Fig.~\ref{fig:decay} shows that the performance of I3D degrades greatly
when the input frame number decreases from $40$, to $20$, $10$ and $5$ on the standard test set of Kinetics and UCF101 split 1. There exists a plenty of redundancy between video frames, and some frames that irrelevant to actions themselves, as shown in Fig.~\ref{fig:sample}. We believe the low performance of I3D on few frame action recognition results from the missing of some key frames that most related to actions. As shown in Fig.~\ref{fig:decay}, the action recognition performance degradation has been alleviated considerably when feeding I3D with distilled frames from TSD.

This work has three main \textbf{contributions}.

\begin{itemize}
    \item We conduct a thorough investigation on the trade-off between the number of frames processed and the accuracy of video action recognition with 3D CNNs, and define the problem of few-frame action recognition.
    \item We explore various approaches of few-frame action recognition. Naive sampling approach (\ie, random sampling and uniform sampling), 
        and supervised approach (\ie, attention based method, TSD) have been implemented and fairly compared.
    \item We propose Temporal Sequence Distillation (TSD), a novel technique that can compress long video clips into compact short ones.  Together with traditional 3D CNNs, it can be jointly trained end-to-end. TSD+I3D typically requires only $50\%$ of the number of frames compared with its counterpart I3D, to achieve almost the same action recognition accuracies,  demonstrating that TSD is very applicable in VA SaaS scenarios.
\end{itemize}

\section{Related work}

In this section, we introduce associated work with our proposed Temporal Sequence Distillation framework. We categorize the related research into four groups: action recognition, temporal-spatial investigation, video summarization and video compression.

\textbf{Action recognition.} Human action recognition developed rapidly in recent years. Especially boosted with several competitive challenges providing public large-scale video data,  deep learning based methods have much overwhelmed previous hand-crafted feature based methods~\cite{LSVC2017, heilbron2015activitynet,kay2017kinetics,karpathy2014large}. Early work of action recognition focuses on designing local hand-crafted features generalized for video, such as Space-Time Interest Points (STIP)~\cite{laptev2005space}, Histogram of 3D Gradient (HOG3D)~\cite{oreifej2013hon4d}, Dense Trajectory (DT)~\cite{wang2013dense}, improved Dense Trajectory (iDT)~\cite{wang2013action}, Motion Boundary Histograms (MBH)~\cite{dalal2006human} and Optical Flow Guided Feature (OFF)~\cite{sun2017optical}. In order to describe frame-level or clip-level video content with a fixed length feature, feature encoding is utilized to aggregate local features. Typical solutions include histogram, bag of words, fisher vector, and vector of locally aggregated descriptors (VLAD)~\cite{jegou2010aggregating}. These multi-step approaches are highly complicated, and errors occurred in the earlier steps are easily accumulated in the later steps. Therefore, their final performance heavily depends on sophisticated design of each step. Two-stream CNN solution brought a significant breakthrough in action recognition~\cite{simonyan2014two}, because deep features pre-trained on ImageNet own better semantic description ability compared with local features. In addition, this deep learning based approach unifies feature extraction,  encoding, and classification in a deep neural network trained in an end-to-end fashion. A temporal segmented two-stream network (TSN) can utilize several video segments simultaneously for action classification during training, which achieved the state of the art performance on several databases~\cite{wang2016temporal}. Encoding frame-level deep features for a video sequence presentation is useful for the two-stream approach. And skip-links between the RGB and optical flow streams can also increase performance~\cite{miech2017learnable,diba2017deep,feichtenhofer2016spatiotemporal}.

\textbf{Temporal-spatial investigation.} Though the classic two-stream solution achieved good performance on action recognition, it relies heavily on optical flow calculation, which is a very time-consuming process. There are some fast-implemented temporal-change estimations for video, which are good replacements of optical flow. Zhu \etal {} proposed MotionNet which takes consecutive video frames as input and estimates motion implicitly accompanying the RGB stream~\cite{zhu2017hidden}. In TSN work, RGB-difference has been investigated to replace optical flow as input for the optical flow stream~\cite{wang2016temporal}, and RGB-difference can obtain a comparable performance with optical flow. The fast implementation of motion estimation can speed up the two-stream approach, however, two separate neural networks are required by the two streams. Since both appearance and motion information are contained in a video sequence, it is a natural choice to utilize only one 3D CNN (C3D) network to explore spatial-temporal information. The first proposal of C3D for action recognition~\cite{tran2015learning} did not perform better than the two-stream solution. With an inflated initialization of C3D (I3D) from 2D CNN (C2D) model pre-trained using ImageNet data, I3D achieved the state-of-the-art performance on UCF101 and Kinetics~\cite{carreira2017quo}. Even using only one RGB stream, I3D can outperform the two-stream approach on kinetics. Due to temporal convolutional operations between frames, I3D is slower than a counterpart C2D with the same depth. Pseudo-3D (P3D) and spatiotemporal-separable 3D (S3D) simulate $3\times 3\times 3$ convolutions with $1\times 3\times 3$ convolutional filters on spatial domain and $3\times 1\times 1$  convolutions to construct temporal connections, thus the computation load of I3D can be reduced by 1.5$\times$. Notice that, typical modalities as optical flow and RGB-difference can model the temporal connections of frames but at the expense of even more redundant computations, which is different from the proposed TSD.

\textbf{Video summarization.}
Potapov \etal~\cite{potapov2014category} utilized Support Vector Machines (SVMs) to predict one importance score for every video shot. After that, video shots with the highest scores were assembled as a sequence to generate a summary. Gong \etal~\cite{gong2014diverse} applied sequential DDP to select diverse subsets, and feature embedding was learned with the guide of manual summary labels. In the work above, hand-crafted features were explored for representing videos shots. Recently, many deep neural networks based video summarization approaches~\cite{kaufman2017temporal,zhang2016summary,zhang2016video,feng2018summarizer} were proposed, which achieved the state of the art performance on public databases~\cite{gygli2014creating,song2015tvsum}. The proposed TSD can be considered as one kind of summarization of videos. However, video summarization approaches focus on extracting frames with content highlight and visual aesthetics. Instead, the proposed TSD targets at maximizing action recognition accuracy with little transmission bandwidth and computation cost by distilling compact frames.

\textbf{Video compression.} High-Efficiency Video Coding (HEVC) is the newest video compression standard~\cite{sullivan2012overview}. Its efficiency mainly relies on accurate frame prediction, including intra-prediction and inter-prediction, which explore spatial redundancy within a single frame, and temporal redundancy between inter-frame respectively. Videos are encoded with only key frames, motion vectors describing block-wise motion between consecutive frames, and residual errors.
Although video compression can reduce transmission bandwidth in VA SaaS applications, different from the proposed TSD, it cannot reduce the computation cost on cloud since the number of frames fed into 3D CNNs (\eg, I3D) keeps unchanged. Moreover, our proposed TSD is orthogonal to video compression approaches, and can be combined with them for further transmission bandwidth reduction.


\section{The Proposed Approach}
In this section,  we first define the problem of few-frame action recognition,  then introduce the proposed TSD in detail,  and finally discuss its relations to other approaches.
\subsection{Few-Frame Action Recognition}

Few-frame action recognition recognizes action based on only a few frames of videos. 
It not only reduces the input data size which can save transmission bandwidth in VA SaaS scenarios, but also speeds up the recognition procedure. Traditional action recognition acceleration approaches explore reducing computation complexity in the dimension of spatial size~\cite{howard2017mobilenets, yue2018boosting, wang2017appearance}, or network architecture~\cite{xie2017rethinking, qiu2017learning}. Few-frame action recognition attempts to accelerate inference from another dimension, \ie, the number of frames fed into networks.

\subsection{Temporal Sequence Distillation}


Towards few-frame action recognition problem, we present \textit{Temporal Sequence Distillation}  to transform a long video sequence of length $T$ into a short one of length $T_s$ ($T_s<T$) automatically.
In order to achieve compactness and informativeness, each distilled frame should be generated according to the global context of the long video sequence. For simplicity, we assume that each distilled frame is a linear combination of all frames of the long sequence. Formally,

\begin{equation}
   Y=XP,
   \label{eq:output}
\end{equation}
where $X$ and $Y$ indicate the frame matrix (with each column as one frame) of the input long sequence and the output short sequence respectively. $P$ is a transformation matrix of shape $T \times T_s$. Its element $P_{ij}$ indicates the importance of the $i_{th}$ frame of the input long video for  the $j_{th}$ distilled frame. Ideally, $P_{ij}$ is nonzero if the $i_{th}$  frame of the input video is relevant to the semantic action category encoded by the input video  otherwise $0$. Herein, we propose a temporal sequence distillation block to predict the transformation matrix $P$ based on input videos.


\subsubsection{Temporal Sequence Distillation Block}

\begin{figure}
\centering
\includegraphics[width=1.0\linewidth]{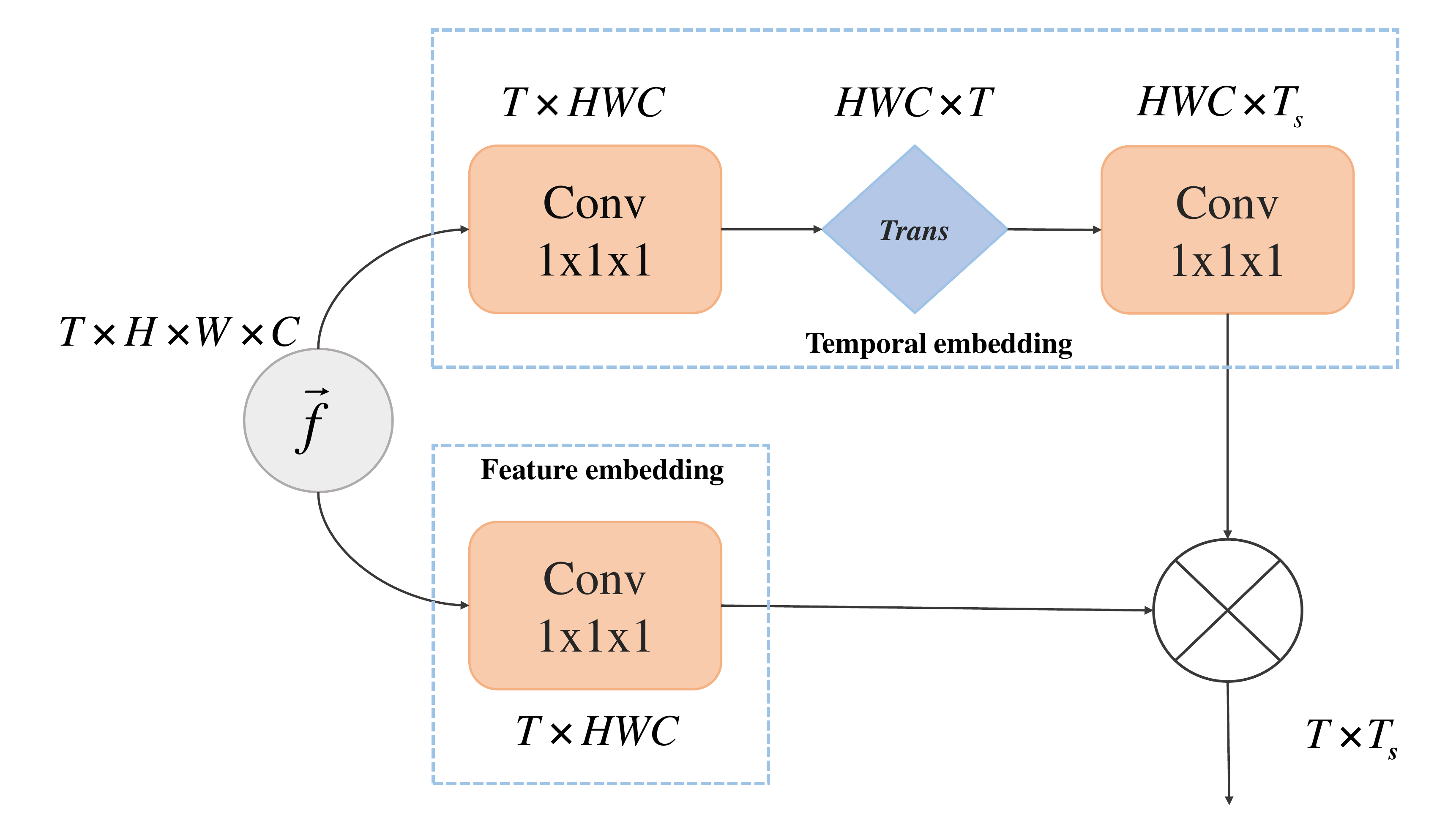}
\caption{Illustration of TSD block.  $\vec{f}$ indicates the input feature of TSD block. $T$, $C$, $H$ and $W$ indicate the input frame number, the channel number,
the feature map height and the feature map width respectively. ${T_s}$ indicates the output frame number of TSD block. The $Trans$ unit  indicates transpose operation, while $\otimes$ represents matrix  multiplication. }
\label{fig:tsd_block}
\end{figure}

Fig.~\ref{fig:tsd_block} shows the architecture of TSD block.
It inputs feature maps of all frames of the input video sequence, and outputs a transformation matrix. 
Given the input feature maps  $\vec{f}$ with size $T\times H\times W\times C$, it produces the transformation matrix $P$ with  one  dual-path network inspired by the  self-attention~\cite{vaswani2017attention} on sentence embedding extraction. 
The first path is designed to extract temporal embedding. It transforms the input feature maps to one feature map $O$ with size $H\times W\times C\times T_s$ via one transpose operation and two convolution operations. The second is designed to extract feature embedding. It transforms the input to one feature map $G$ with size $T\times H\times W\times C$ via one convolution operation. Formally, we have
\begin{equation}
    O=\text{Trans}(\vec{f}*w_{\alpha})*w_{\beta},
\end{equation}
and
\begin{equation}
    G=\vec{f}*w_{\gamma},
\end{equation}
where $*$ indicates convolution operation, and $\text{Trans}()$ indicates transpose from a tensor with size $T\times H\times W\times C$ to one with size $H\times W\times C\times T$. $w_{\alpha}$,$w_{\beta}$ and $w_{\gamma}$ are 3D convolution kernels.

The transformation matrix $P$ is given by
\begin{equation}
    P=\text{softmax}(O^{'}G^{'}),
\end{equation}
where $O^{'}$ and $G^{'}$ are two matrices of size $HWC\times T_s$ and $T\times HWC$, which are reshaped from $O$ and $G$ respectively. $\text{softmax}()$ column-wisely normalizes an input matrix by softmax.





%
\begin{figure*}
\centering
\includegraphics[width=1.0\linewidth]{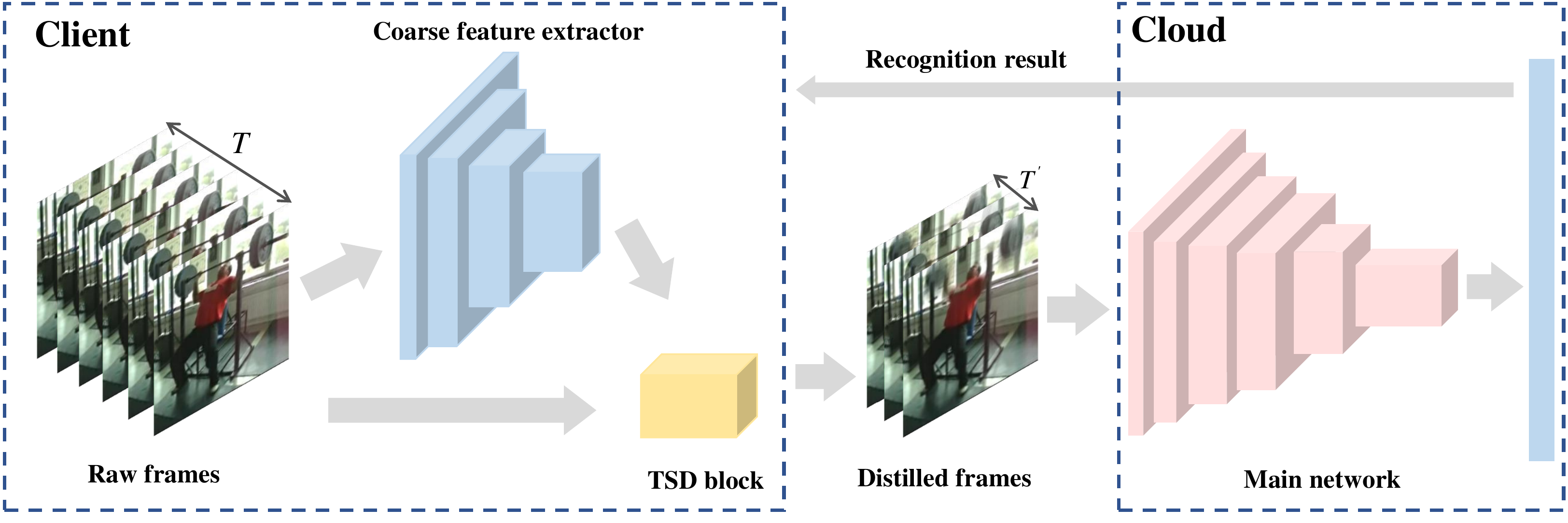}
\caption{Pipeline of TSD network in VA SaaS scenarios. Only the distilled frames are transmitted.}
\label{fig:tsd_net}
\end{figure*}
\subsubsection{Temporal Sequence Distillation Network}

The proposed Temporal Sequence Distillation network (TSD network) consists of
three components. Namely, the coarse feature extractor, the TSD
block, and the main network. Fig. 5 visualizes its pipeline. The feature
extractor is designed to extract coarse feature of the video
snippets $X$. The successive TSD block takes the coarse feature as
input, and predicts the corresponding transformation matrix $P$. The
distilled short sequence $Y$ is calculated as Equation~\ref{eq:output}. The main
network is a CNN with temporal convolution to recognize video
action. In this paper, we take I3D (Inception v1 with BN) as the
main network. The pipeline is entirely end-to-end, and could be trained and evaluated jointly. During evaluation, the TSD network
could be separated into two components: coarse feature extractor+TSD
block, named client component, and main network, named cloud
component. The two components are mutually parameter independent,
thus they can be deployed onto terminals and cloud
separably. In this way, only the distilled frames are transmitted
to cloud to
carry out the video recognition task. This can accelerate the data
transmission and relief the bandwidth pressure.


\subsubsection{Coarse Feature Extractor}
We design the coarse feature extractor for TSD network with keeping efficiency in mind.
We choose the current popular light-weight  CNN MobileNet~\cite{howard2017mobilenets} as the feature extractor. For acceleration, we downsample the input of the coarse feature extractor from $224\times 224$ to  $112\times112$.



\subsection{Relations to Other Approaches}

Sampling a subset of input frames is one natural way to distill long video sequences into short ones.
There exist two naive sampling strategies, namely, random sampling and uniform sampling which are widely used in other methods such as TSN~\cite{wang2016temporal}.

\textbf{Random Sampling}.
A subset of
input frames are randomly sampled. We denote I3D with the random sampling when training and test as I3D$_{R}$.

\textbf{Uniform Sampling}. A subset of input frames with equal intervals are uniformly sampled. We denote I3D with uniform sampling when training and test as I3D$_{E}$. 

\textbf{Temporal Attention.}
During training, we learn a scalar importance weight  for each input frame.
All input frames scaled with their corresponding importance weight are fed into the main action recognition network (I3D)  for end-to-end training.
During testing, we select frames with the top $T_s$ predicted importance, which are fed into the main network for action recognition.
We denote I3D with temporal attention as I3D$_{A}$.

All the above methods follow the same pipeline of frame selection as defined in Equation~\ref{eq:output}.
The differences are how to generate the transformation matrix $P$. For random and uniform sampling, it randomly or uniformly activates one value at each column. For temporal attention, it learns one $l1$ normalized importance weight for each column. 


\section{Experiments}
\label{sec:exp}
We conduct comprehensive ablation experiments along with a thorough comparison between the proposed TSD and the state-of-the-art action recognition methods.
\subsection{Datasets and Evaluation}
We evaluate the proposed TSD on two popular datasets, namely, UCF101 and Kinetics. Both UCF101 and Kinetics contain short trimmed videos, each of which is annotated with one action label. UCF101 consists of $13320$ videos from $101$ classes and has three splits in the testing set. We report the average performance of $3$ splits except as otherwise noted. Kinetics contains $240000$ training videos from $400$ human action classes with $400$ or more samples for each class, and $40000$ testing videos with 100 for each class. All the videos are trimmed and last around $10$s.

During testing, we randomly sample $Q$ times and report the average of probabilities as final prediction. We fairly compare the proposed TSD with its counterparts with the same sampling times. We also conduct experiments with different $Q$ (\ie, 1, 3, and 10) to study the effect of the sampling times of the proposed TSD.

The evaluation procedures of few frame action recognition are as follows:
\begin{itemize}
\item \textbf{I3D}.  It is trained as done in~\cite{carreira2017quo} with randomly sampled consecutive 64-frame clips. During testing, we sample consecutive $T_{s}$ frames to evaluate and then average predictions temporally as done in~\cite{carreira2017quo}. We extensively
    evaluate its performance on few frame action recognition by setting $T_{s}$ to $5$, $10$, $20$, and $40$.
\item \textbf{I3D$_R$}.   We randomly sample $T$-frame clips, from each of which $T_{s}$ ($T\ge T_{s}$) \textit{ordered} frames are randomly sampled and fed into $I3D$ during both training and testing.
\item \textbf{I3D$_E$}.   We randomly sample $T$-frame clips, and then sample $T_{s}$ frames with equal intervals as input from each clip during both training and testing.
\item \textbf{I3D$_A$}.   We randomly sample $T$-frame clips, which are fed into the I3D network with temporal attention during training. $T_{s}$ frames with the top $T_{s}$ importance scores are selected from each clip for action recognition during testing.
\item \textbf{I3D+TSD}.   We randomly sample $T$-frame clips, which are fed into the TSD network, and then take the outputted distilled clip of length $T_{s}$ as input to the I3D network.
\end{itemize}





\setlength{\tabcolsep}{8pt}
\begin{table*}[t]
 \centering
 \caption{\textbf{Comparison with baseline and other frame selection approaches on Kinetics. All are tested on RGB modality with Inception v1 as backbone and with 3-clip evaluation ($Q=3$) for fair comparison. $\Delta$ indicates the performance improvement compared with the baseline I3D.}
 }

 \begin{tabular}{c|c|c|c|c|c|c}
\hline $\#$sampled frames ($T_{s}$) & Clip length ($T$) & I3D &I3D$_A$/$\Delta$ & I3D$_R$/$\Delta$  & I3D$_E$/$\Delta$  & I3D+TSD/$\Delta$  \\ \hline

5 & 5 & 36.7 & 34.6 /-2.1& - & - & - \\
 & 10 & - & 35.6/-1.1 & 37.1/+0.4 & 39.9/+3.2 & 65.7/+29.0 \\
& 20 & - & 36.1/-0.6 & 40.9/+4.2 & 41.7/+5.0 &\textbf{ 65.9/+29.2 }\\ \hline

 10 & 10 & 65.7 & 63.0/-2.7 & -& - & - \\
 & 20 & - & 64.1/-1.6 & 65.8/+0.1 & 66.1/+0.4 & 67.9/+2.2 \\
& 40 & - & 65.2/-0.5 & 65.9/+0.2 & 66.5/+0.8 & \textbf{68.6/+2.9} \\ \hline

20 & 20 & 68.4 & 68.6/+0.2 & - & - & - \\
& 40 & - & 68.8/+0.2 & 69.0/+0.6 & 69.2/+0.8 & 69.9/+1.5 \\
& 80 & - & 68.8/+0.2 & 68.6/+0.2 & 69.9/+1.5 &\textbf{ 70.7/+2.3} \\ \hline

 40 & 40 & 71.0 & 71.1/+0.1 & - & - & - \\
& 80 & - & 71.4/+0.4 & 70.8/-0.2 & 71.4/+0.3 & \textbf{72.4/+1.4} \\ \hline

 \end{tabular}
 \label{tab:tsd}
\end{table*}

\setlength{\tabcolsep}{2pt}
\begin{table}
\small
\begin{center}
\caption{Effect of the clip number ($Q$). }
\label{tab:clip}
\begin{tabular}{lllll}
\hline\hline\noalign{\smallskip}
 $\#$Clips (Q) & Method &  Clip length ($T$) & $\#$sampled frames ($T_{s}$)& Accuracy \\
\noalign{\smallskip}
\hline
\noalign{\smallskip}
1& I3D & 5 & 5 & 34.7\\
& I3D+TSD& 20 & 5 & 58.0 \\

& I3D& 10  & 10 & 57.9 \\
& I3D+TSD  & 40 & 10 & 62.2 \\
& I3D & 20 & 20 & 61.9 \\
& I3D+TSD & 80 & 20 & 64.6 \\
& I3D & 40 & 40 & 64.7 \\
& I3D+TSD & 80 & 40 & \textbf{65.9} \\
\hline
3& I3D& 5 & 5 & 36.7\\
& I3D+TSD & 20 & 5 & 65.9 \\
& I3D& 10 & 10 &65.7 \\
& I3D+TSD & 40 & 10 & 68.6 \\
& I3D & 20 & 20 & 68.4 \\
& I3D+TSD & 80 & 20 & 70.7 \\
& I3D & 40 & 40 & 71.0 \\
& I3D+TSD & 80 & 40 & \textbf{72.4 }\\
\hline
10& I3D & 5 & 5 & 37.4 \\
& I3D+TSD  & 20 & 5 & 67.8 \\
& I3D& 10 & 10 & 67.9 \\
& I3D+TSD & 40 & 10 & 70.2 \\
& I3D & 20 & 20 & 70.6 \\
& I3D+TSD  & 80 & 20 & 72.1 \\
& I3D & 40 & 40 & 71.8 \\
& I3D+TSD  & 80 & 40 & \textbf{73.2} \\ \hline \hline

\end{tabular}
\end{center}
\end{table}
\setlength{\tabcolsep}{1.4pt}

\subsection{Implementation Details}
All of our experiments are done using Tensorflow on one cloud with 8 Titan X GPU servers.

\textbf{Training I3D.} We train the I3D model by following the practice in~\cite{carreira2017quo}.

\textbf{Training TSD on Kinetics.} We initialize the coarse feature extractor with the MobileNet model
pretrained on ImageNet,
and its sequence distillation module using Gaussian with standard variance $0.01$. The main branch is initialized with the pre-trained model on Kinetics.
We train TSD in two stages on Kinetics. First, we train the coarse feature extractor and the TSD block  while fixing the main branch. We employ multi-step learning rate policy with the base learning rate being 0.1. The learning rate decays by 10 times every 10000 iterations, and the maximum iteration size is set to $50000$. Second, we train the whole TSN including the coarse feature extractor, the TSD block and the main branch end-to-end. We set the learning rate to $0.01$ with the decay factor of $0.3$ and the decay step of $30000$.

\textbf{Training TSD on UCF101.}
  All the parameters of the TSD network other than the final FC layer are first initialized with the pretrained weights on Kinetics, and then the whole TSD network is trained on UCF101 end-to-end. Note that the learning rate of the final FC layer is set to  10 times of those of other layers.

\subsection{Effectiveness of TSD}
\subsubsection{Comparison of Sampling Methods}
We compare the proposed I3D+TSD with its baseline I3D and other sampling approaches in Table~\ref{tab:tsd}. It has been shown that all sampling methods outperform the baseline I3D. We believe that this is because of these sampling methods keeping the same frame selection procedure during both training and test. Among all sampling methods, the proposed I3D+TSD achieves the biggest performance improvement. Specifically, I3D+TSD improves the accuracy of I3D on Kinetics by $29.2$, $2.9$, $2.4$, and $1.4$ when the sampled frame number $T_{s}$ is set to 5,10,20, and 40 respectively.

\subsubsection{Effect of Input Clip Number ($Q$)}
Table \ref{tab:clip} investigates the effect of the input clip number.
For each clip, we generate a short sequence of frames with TSD which are fed into I3D, achieving category probabilities. The averaged probabilities over all clips are considered as the final prediction. It has been shown that  I3D+TSD always outperforms its counterpart I3D regardless of the sampling clip number. Increasing the input clip number can improve action recognition accuracies for both I3D+TSD and I3D.

\setlength{\tabcolsep}{4pt}
\begin{table}
\small
\begin{center}
\caption{Comparison between I3D and I3D+TSD on UCF101 split1. All are 1-clip evaluated on RGB modality with Inception v1 with BN as backbone for fair comparison. }
\label{table:ucf}
\begin{tabular}{lllll}
\hline\noalign{\smallskip}
 \#sampled frames ($T_{s}$) & Method & Clip length ($T$) & Accuracy \\
\noalign{\smallskip}
\hline
\noalign{\smallskip}
5 &I3D& 5  & 79.7\\
 &I3D+TSD & 10  & 84.9 \\
&I3D+TSD & 20  &\textbf{87.4}\\ \hline

10&I3D& 10 & 87.7\\
&I3D+TSD & 20 & 89.4  \\
&I3D+TSD & 40 & \textbf{90.6}\\
 \hline
20&I3D& 20 & 90.9\\
&I3D + TSD  & 20 & 92.3 \\
&I3D + TSD  & 20 & \textbf{92.7}  \\  \hline

40&I3D & 40  & 93.3  \\
&I3D+TSD & 80 & \textbf{94.2} \\ \hline \hline
\end{tabular}
\end{center}
\end{table}
\setlength{\tabcolsep}{1.4pt}

\subsubsection{Experiments on Other Dataset}
We further validate the effectiveness of TSD on the classic UCF101 dataset in Table~\ref{table:ucf}. Similarly,  I3D+TSD achieves performance improvements as on Kinetics. Specifically, it outperforms I3D by 7.7, 2.9, 1.8, and 0.9 when the sampled frame number $T_{s}$ is set to 5, 10, 20, and 40 respectively. I3D+TSD performs much better than I3D on few frame action recognition. Therefore, I3D+TSD is more applicable than I3D in VA SaaS scenarios. 

\subsection{Efficiency}
TSD can be considered as one way of network acceleration. We compare parameter numbers and FLOPS of the proposed TSD+I3D and its counterpart I3D with comparable accuracy in Table~\ref{table:speed}. All the speed is measured on servers with 8 Titan X GPUs. Although I3D+TSD has more parameters, however, it has about $32\%\sim 33\%$ less FLOPS, and is about $13\%\sim 26\%$ faster than I3D.  When I3D+TSD is deployed in VA SaaS scenarios, it has even about $50\%\sim 51\%$ less FLOPS, and is about $45\%\sim 48\%$ faster than I3D on cloud.

\setlength{\tabcolsep}{4pt}
\begin{table}
\small
\begin{center}
\caption{Efficiency comparison. All are tested on Kinetics with RGB modality and 3 clips ($Q=3$). The I3D$_{(T)}$ represents  I3D with length-\textbf{$T$} clips. The I3D+TSD$_{(T\rightarrow T_{s})}$ represents the I3D fed with \textbf{$T_{s}$} distilled  frames from  length-\textbf{$T$}  clips in original videos. The I3D+TSD$_ {(T\rightarrow T_{s})}^{cloud}$ indicates I3D+TSD is deployed in VA SaaS scenarios (\ie, only the main network is deployed on cloud while others on client terminals). $\Delta$ indicates FLOPS or speed improvement compared with baselines.}
\label{table:speed}
\begin{tabular}{lllll}
\hline\noalign{\smallskip}
Method & Accuracy &\tabincell{c}{\#params\\(M)} & \tabincell{c}{FLOPS /$\Delta$ \\(G)}&\tabincell{c}{Speed/$\Delta$ \\(ms)}  \\
\noalign{\smallskip}
\hline
\noalign{\smallskip}
I3D$_{(10)}$& 65.7 & 12.1 & 51 & 46.1 \\
I3D+TSD$_{(20\rightarrow5)}$ & 65.9 & 16.4 & 34/-17 & \textbf{39.9/-6.2} \\
I3D+TSD$^{cloud}_{(20\rightarrow 5)}$& 65.9 & 12.1 & 25/-26 & \textbf{23.9/-22.2 }\\ \hline

I3D$_{(20)}$& 68.4 & 12.1 & 101 & 77.9\\
I3D+TSD$_{(40\rightarrow10)}$ & 68.6 & 16.4 & 69/-32 &\textbf{ 58.2/-19.7} \\
I3D+TSD$^{cloud}_{(40\rightarrow 10)}$& 68.6 & 12.1 & 51/-50 & \textbf{43.1/-34.8} \\ \hline

I3D$_{(40)}$& 71.0 & 12.1 & 202 & 150.3 \\
I3D+TSD$_{(80\rightarrow20)}$ & 70.7 & 16.4 & 137/-65 &  \textbf{111.0/-39.3}\\
I3D+TSD$^{cloud}_{(80\rightarrow20)}$ & 70.7 & 12.1 & 101/-101 & \textbf{77.9/-72.4} \\ \hline

\end{tabular}
\end{center}
\end{table}
\setlength{\tabcolsep}{1.4pt}

\section{Conclusions}
We have proposed a novel Temporal Sequence Distillation technique for few frame action recognition by distilling long video sequences into short compact sequences. It consists of coarse feature extractor, temporal sequence distillation block, and main network (\eg, I3D), all of which can be trained in an end to end way. Because it distills key frames relevant to actions and discards considerable temporal redundancy, it can accelerate action recognition while preserving recognition accuracy.  When deploying its coarse feature extractor and TSD block on client terminals, and its main network on cloud in VA SaaS scenarios, it can greatly reduce the data transmission bandwidth between client and cloud.


\begin{acks}
This work is partially supported by SenseTime Group Limited, the Hong Kong Innovation and Technology Support Programme, and the National Natural Science Foundation of China (61503366).
\end{acks}

\bibliographystyle{ACM-Reference-Format}
\bibliography{references}

\end{document}